\documentclass{article}
\usepackage{ijcai26}

\usepackage{times}
\usepackage{soul}
\usepackage{url}
\usepackage[hidelinks]{hyperref}
\usepackage[utf8]{inputenc}
\usepackage[small]{caption}
\usepackage{graphicx}
\usepackage{amsmath}
\usepackage{amsthm}
\usepackage{subfig}
\usepackage{booktabs}
\usepackage{algorithm}
\usepackage{algorithmic}
\usepackage[switch]{lineno}
\usepackage{amssymb}
\usepackage{multirow}
\usepackage{color, colortbl}
\usepackage{orcidlink}
\usepackage{booktabs}
\definecolor{Worse}{rgb}{0.9,0.9,1}
\definecolor{Better}{rgb}{1,0.9,0.9}

\title{Discrete Cosine Transform-Based Decorrelated Attention for Vision Transformers
}

\author{
Hongyi Pan$^1$
\and
Emadeldeen Hamdan$^2$
\and
Xin Zhu$^2$
\and
Ahmet Enis Cetin$^2$
\and
Ulas Bagci$^1$\\
\affiliations
$^1$Machine and Hybrid Intelligence Lab, Northwestern University, Chicago, USA\\
$^2$Department of Electrical and Computer Engineering, University of Illinois Chicago, Chicago, USA\\
\emails
\{hongyi.pan, ulas.bagci\}@northwestern.edu, \{ehamda3, xzhu61, aecyy\}@uic.edu, 
}

\begin{document}

\maketitle

\begin{abstract}
Self-attention is central to the success of Transformer architectures; however, learning the query, key, and value projections from random initialization remains challenging and computationally expensive. In this paper, we propose two complementary methods that leverage the Discrete Cosine Transform (DCT) to enhance the efficiency and performance of Vision Transformers. First, we address the initialization problem by introducing a simple yet effective DCT-based initialization strategy for self-attention, where projection weights are initialized using DCT coefficients. This structure-preserving approach consistently improves classification accuracy on the CIFAR-10 and ImageNet-1K benchmarks. Second, we propose a DCT-based attention compression technique that exploits the decorrelation properties of the frequency domain. By observing that high-frequency DCT coefficients typically correspond to noise, we truncate high-frequency components of the input patches, thereby reducing the dimensionality of the query, key, and value projections without sacrificing accuracy. Experiments on Swin Transformer models demonstrate that the proposed compression method achieves a substantial reduction in computational overhead while maintaining comparable performance. Code: \url{https://github.com/NUBagciLab/DCT-Transformer}.
\end{abstract}

\section{Introduction}
In the last decade, convolutional neural networks (CNNs) have dominated image-processing tasks. However, inspired by the success of transformer architectures in natural language processing, Vision Transformers (ViTs) have redefined the norms of visual data interpretation in recent years. Self-attention mechanism, a key component of transformers, is pivotal in reshaping how models perceive and analyze visual information~\cite{vaswani2017attention}. 

Unlike CNNs, ViTs view an image as a sequence of informative patches, adopting a more holistic approach. By treating image patches like ``visual words'', ViTs can leverage powerful transformer architectures, originally designed for natural language processing, to learn global context and long-range dependencies. This novel approach has led to remarkable performance gains on a wide range of computer vision tasks, demonstrating the potential of ViTs to revolutionize the field.

The initialization of self-attention weights is a critical step significantly impacting the training dynamics and performance of the Transformers. The core of Transformers is the \textit{self-attention} mechanism, which includes three weight matrices: queries ($\mathbf{W}_Q$), keys ($\mathbf{W}_K$), and values ($\mathbf{W}_V$). Proper initialization of these matrices ensures that the model starts from a stable state, facilitating effective learning. The most common approach is to initialize the weights randomly, often using distributions like the normal or uniform distribution. For example, Xavier initialization~\cite{glorot2010understanding} and He initialization~\cite{he2015delving} are popular choices. They are designed to keep the scale of the gradients roughly the same in all layers. Although less common, in some cases, biases or certain parts of the weight matrices are initialized to zeros~\cite{masood2012training,zhao2022zero}. This is less common for queries, keys, and values matrices, as it can lead to symmetry-breaking issues during training~\cite{hu2019exploring}.

\textbf{Efficiency is an ongoing concern in Transformers.} While conventional initialization strategies are often suitable for generic applications, new algorithms are increasingly tailored to the specific characteristics and needs of Transformers. These advancements in initialization techniques contribute to the ongoing improvements in the performance and efficiency of Transformer-based models. Orthogonal initialization~\cite{saxe2013exact,hu2019provable}, sparse initialization~\cite{martens2010deep}, variance scaling~\cite{zhang2019improving}, pre-trained initialization~\cite{weiss2016survey,bozinovski2020reminder,xu2023initializing}, layer-wise, domain-specific and attention-specific initialization~\cite{trockman2023mimetic} are some of the new trends in initialization approaches. In this study, we propose a new family of algorithms for initialization strategy in the intersection of attention and domain-specific approaches: \textit{Discrete Cosine Transform-Based Decorrelated Attention}.

\textbf{Can DCT in Vision Transformer be a game changer?} DCT has been widely used in signal processing and data compression applications. It reduces the dimensionality of the input data while preserving its essential features. Transforming the input data into the frequency domain helps capture the most important features of the input data. In this paper, we propose two methods that apply DCT to the attention for vision transformers. First, we initialize the weight matrices for the queries, keys, and values computation as the DCT matrix. Our DCT-based initialization covers the entire spectrum utilizing the DCT basis vectors, and each weight vector has a different initial bandwidth. As a comparison, to cover the entire spectrum in the classical initialization, all the weight matrices are initialized as random numbers due to their independent generation. These random numbers can be treated as white noise. All the weight initialization starts with the same spectra. Second, we introduce a DCT-based compressed attention for vision transformers. With our compression methodology, compared to the vanilla Swin Transformers, the number of parameters and computational overhead of the Swin Transformers are reduced. Meanwhile, the accuracy of the models is comparable.

\section{Related Works}
\paragraph{Vision Transformer}  
Transformer architectures have merged as a dominant standard for natural language processing tasks~\cite{vaswani2017attention}. 
Vision Transformer (ViT) \cite{dosovitskiy2020image} built a bridge between language and vision. It shows that pure transformers can also be applied directly to sequences of image patches for classification tasks. In other words, it partitions images into patches and treats them as word sequences. Then, a shifted windowing scheme was proposed in Swin Transformer~\cite{liu2021swin} to obtain greater efficiency. This limits self-attention computation to non-overlapping local windows while allowing for cross-window connection. Later, to overcome the limitations of the heavyweights in ViTs, MobileViT~\cite{mehta2021mobilevit} combined the strengths of MobileNets~\cite{sandler2018mobilenetv2} and ViTs. It is a lightweight and low-latency network for mobile vision tasks. Moreover, gSwin \cite{go2023gswin} combined MLP-based architecture with Swin Transformer and attains superior accuracy with a smaller model size. 

\paragraph{Frequency-based Neural Network} 
Orthogonal transforms such as the Fourier transform~\cite{chi2020fast,buchholz2022fourier,patro2025spectformer} and Hadamard transform~\cite{pan2023hybrid} 
have been used to improve the performance of the neural networks. In the orthogonal transform family, DCT is a mathematical technique commonly used in signal processing and image and video compression. 
Compared to the Fourier transform, which is the standard frequency representation, DCT does not generate complex-valued numbers. Therefore, it is computationally more efficient. DCT-Former~\cite{scribano2023dct}, for instance, applied DCT compression on the input sequence along the length channel before feeding it into the self-attention layers to reduce the computational overhead in the NLP problems. Furthermore, inspired by JPEG compression, DCFormer~\cite{li2023discrete} highlighted the visually significant signals by leveraging the input information compression on its frequency domain representation. These two DCT-based works and our DCT-based compression all aim to reduce the computational overhead. Therefore, we mainly compare our DCT-based compression method with them. DCT-Former~\cite{scribano2023dct} is designed for the NLP transformers and takes DCT along the width and height of the input tensor patches, while our method is designed for the vision transformers and it compresses the tensor along the channel axis. DCFormer~\cite{li2023discrete} averages DCT coefficients via average pooling, which does not take advantage of the property that DCT components in low-frequency bands contain more important information. On the contrary, our DCT-based compression method retains the low-frequency components very accurately and discards the unnecessary extremely high-frequency components corresponding to noise. Under the same input image size, compared to the vanilla transformer models, DCFormer reduces the FLOPs and retains the number of parameters while sacrificing accuracy. Our DCT-based compression method, on the other hand, saves both trainable parameters and FLOPs and obtains on-par or better performance in addition to other benefits that our DCT-based compression method has. 
Furthermore, we introduce a DCT-based initialization and we have not found any related initialization in the literature.

\paragraph{Self-Attention Initialization} 
DS-Init~\cite{zhang2019improving} improves convergence in deep Transformers for NLP by reducing parameter variance during initialization and minimizing output variance of residual connections, facilitating better gradient flow. It pairs with a merged attention sublayer (MAtt) that combines average-based self-attention with encoder-decoder attention to reduce the computational cost. Additionally, Mimetic initialization \cite{trockman2023mimetic} is a compute-free method that initializes the weights of self-attention layers by simulating the patterns found in pre-trained weights, resulting in improved Transformer performance. Moreover, weight selection~\cite{xu2023initializing} enables the initialization of smaller models by selecting a subset of weights from a larger pre-trained model, facilitating efficient knowledge transfer and faster training. This method enhances small model performance, reduces training time, and integrates with knowledge distillation, making it particularly valuable for resource-constrained environments. Furthermore, structured initialization~\cite{zheng2024structured} tackles the challenge of training vision transformers on small-scale datasets by introducing a structured initialization strategy inspired by the architectural biases of Convolutional Neural Networks (CNNs). While these methods focus on mimicking pre-trained patterns, selecting weights from larger models, or using CNN-inspired biases, our DCT-based initialization utilizes well-established DCT matrices to initialize attention weights.

\paragraph{Efficient Self-Attention} 
A variety of approaches have been proposed to reduce the memory and computational costs of dot-product attention, including linear-complexity attention mechanisms~\cite{shen2021efficient}, increasing attention heads as in Hydra Attention~\cite{bolya2022hydra}, and architectural rethinking exemplified by MetaFormer~\cite{yu2022metaformer}, which shows that overall architecture often matters more than the specific token mixer. MicroViT~\cite{setyawan2025microvit} further targets resource-constrained settings by introducing an Efficient Single Head Attention mechanism that reduces self-attention cost through group convolution and partial channel processing. Despite these advances, existing methods mainly pursue efficiency through linearization, head scaling, or architectural design, while largely ignoring frequency-domain transformations. We propose DCT-based methods to enhance both efficiency and expressiveness in vision transformers.

\begin{figure}[tb]
\begin{center}
\includegraphics[width=1\linewidth]{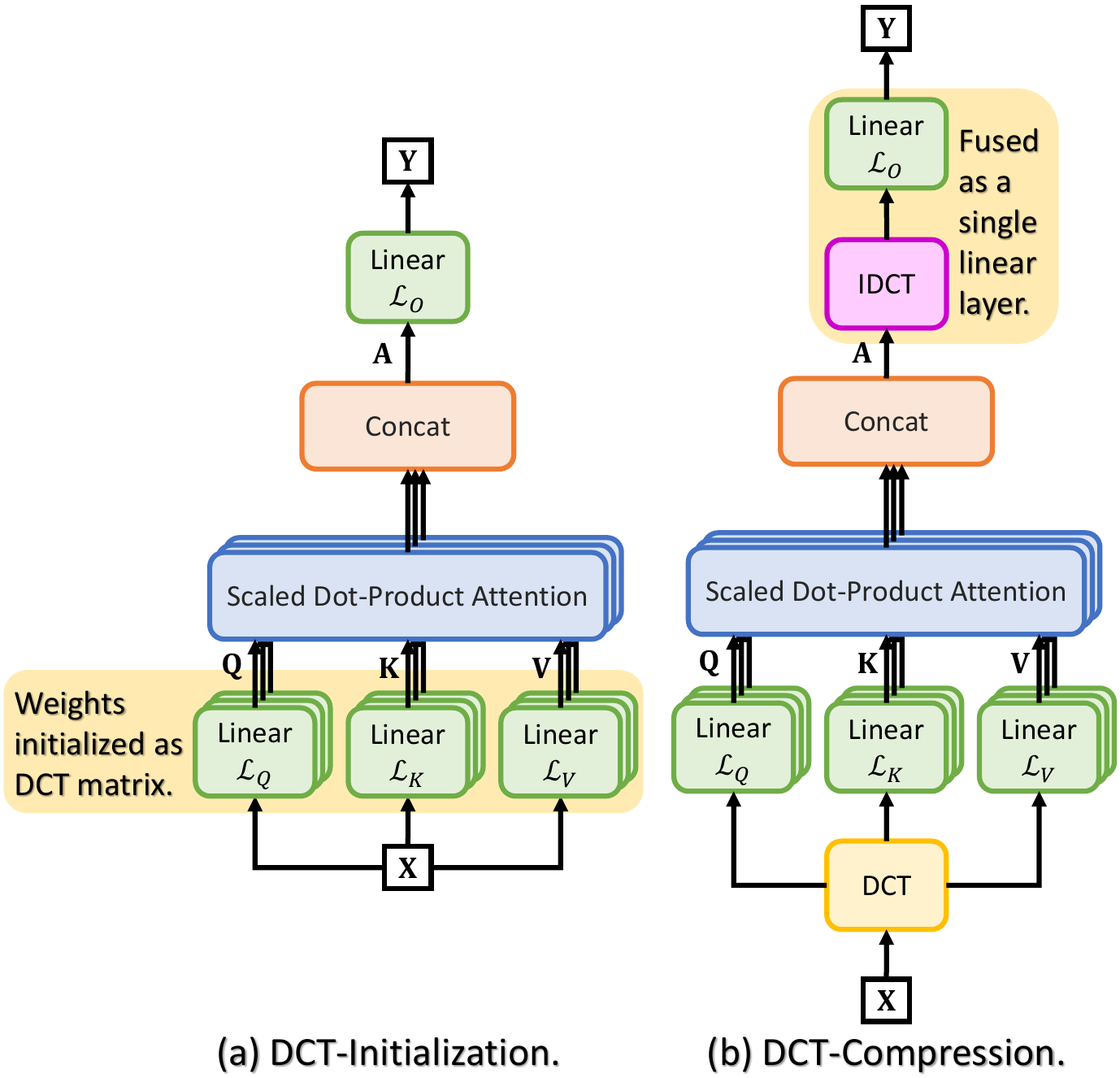}
\caption{DCT-based initialization and compressed MSA.}
\label{fig:MSA}
\end{center}
\end{figure}

 \section{Methodology}
\subsection{Background: Multi-Head Self-Attention}
To handle a 2D image tensor $\mathbf{I}\in \mathbb{R}^{H\times\ W\times C}$, the multi-head self-attention (MSA) module first splits it into patches $\mathbf{X}\in\mathbb{R}^{N\times M^2 \times C}$, where $H\times W$ are the resolution of the image and $C$ is the number of channels, $M\times M$ is the resolution of each image patch and $N=\frac{WH}{M^2}$ is the resulting number of patches. After that, three linear layers ($\mathcal{L}_Q$, $\mathcal{L}_K$, $\mathcal{L}_V$) are applied on $\mathbf{X}$ to compute the queries, keys, and values $\mathbf{Q}, \mathbf{K}, \mathbf{V}\in \mathbb{R}^{N\times M^2 \times C}$:
\begin{equation}
    \mathbf{Q} = \mathbf{XW}_Q^T+\mathbf{b}_Q,
    \mathbf{K} = \mathbf{XW}_K^T+\mathbf{b}_K,
    \mathbf{V} = \mathbf{XW}_V^T+\mathbf{b}_V,
    \label{eq:linear in}
\end{equation}
where $\mathbf{W}_Q, \mathbf{W}_K, \mathbf{W}_V\in\mathbb{R}^{C \times C}$ and $\mathbf{b}_Q, \mathbf{b}_K, \mathbf{b}_V\in \mathbb{R}^{1 \times C}$ are trainable weights and biases. Then, to perform $P$-head MSA, $\mathbf{Q, K, V}$ are split into $P$ slices as $\mathbf{Q}_i, \mathbf{K}_i, \mathbf{V}_i\in \mathbb{R}^{N \times M^2 \times d}$ for $i=0,\ldots, P-1$, $d=\frac{C}{P}$. The scaled dot-product attention $\mathbf{A}_i\in\mathbb{R}^{N\times M^2 \times d}$ is computed as: 
\begin{equation}
    \mathbf{A}_i=\text{Attention}(\mathbf{Q}_i, \mathbf{K}_i, \mathbf{V}_i) = \sigma\left(\frac{\mathbf{Q}_i\mathbf{K}_i^T}{\sqrt{d}}+\mathbf{B}_i\right)\mathbf{V}_i,
    \label{eq:attention}
\end{equation}
where $\sigma(\cdot)$ is the \textit{Softmax} function. $\mathbf{B}_i\in\mathbb{R}^{M^2\times M^2}$ is the relative position bias. Since the relative position along each axis lies in $[-M+1, M-1]$, values in $B_i$ are taken from a parameterized smaller-sized bias matrix $\hat{\mathbf{B}}_i\in \mathbb{R}^{(2M-1)\times (2M-1)}$. After that, $\{\mathbf{A}_i\}$ are concatenated to $\mathbf{A}\in\mathbb{R}^{N\times M^2 \times C}$ to feed into a linear layer $\mathcal{L}_O$ to obtain the attention output $\mathbf{Y}\in\mathbb{R}^{N\times M^2 \times C}$:
\begin{equation}
    \mathbf{Y} = \mathbf{AW}_O^T+\mathbf{b}_O.
    \label{eq:linear out}
\end{equation}
Finally, $\mathbf{Y}$ is reversed into the image shape $\mathbf{J}\in \mathbb{R}^{H\times W\times C}$.

\subsection{DCT-based Initialization for Attention}
In the attention layer, the first three linear layers ($\mathcal{L}_Q$, $\mathcal{L}_K$, and $\mathcal{L}_V$) extract different features of the input patches $\mathbf{X}$. However, their weights $\mathbf{W}_Q$, $\mathbf{W}_K$, and $\mathbf{W}_V$ are trained simultaneously. Such training from scratch is difficult, especially when the dataset is insufficient and small-scale. On the other hand, DCT is a well-designed feature extractor. In our proposed DCT initialization strategy, one of $\mathbf{W}_Q$, $\mathbf{W}_K$, $\mathbf{W}_V\in\mathbb{R}^{C\times C}$ is initialized using the orthogonal type-II DCT matrix $\mathcal{D}$~\cite{ahmed1974discrete}:
\begin{equation}
    \resizebox{.91\linewidth}{!}{$
            \displaystyle
    \{\mathbf{X}\mathcal{D}^T\}[k] = \begin{cases}\frac{\sqrt{2}}{C}\sum_{n=0}^{C-1}\mathbf{X}[n], &k=0,\\
    \frac{2}{C}\sum_{n=0}^{C-1}\mathbf{X}[n]\cos\frac{(2n+1)k\pi}{2nC}, &k=1,\ldots, C-1.\end{cases}
            $}.
\end{equation}
$\mathcal{D}$ can be obtained from applying DCT on an identity matrix $\mathbf{I}_{C\times C}$ because DCT is a linear transform. The remaining weight matrices and other trainable parameters are still initialized regularly. 
We don't initialize multiple weight matrices to avoid the weight symmetry problem~\cite{hu2019exploring} caused by initializing the multiple weights as the same values, and our ablation study has verified this point.

The DCT-based initialization can be treated as applying DCT instead of linear projection on $\textbf{X}$ to obtain $\mathbf{Q}$, $\mathbf{K}$, or $\mathbf{V}$. Some DCT coefficients capture low-frequency information, while others capture high-frequency details. Therefore, the DCT matrix is a good starting point for the weight matrices. 

\subsection{DCT-Based Compressed Attention}
Data compression using DCT is common in signal, image, and video processing for energy concentration, compression efficiency, and noise reduction. 
Usually, higher-frequency DCT coefficients contribute less to perceptual quality and contain more noise. 
Preserving essential information in lower frequencies while discarding less critical details in higher frequencies results in efficient signal representation and storage.
The specific truncation strategy depends on the application's requirements and the acceptable level of information loss.

We observed that the attention layer matrices are correlated with each other, so that we can compress them using DCT.
Fig.~\ref{fig:MSA}(b) shows the method of DCT-based compression for the attention layer. First, the input patches $\mathbf{X}\in\mathbb{R}^{N\times M^2 \times C}$ are encoded by DCT along the channel of $C$ with the high-frequency components truncated. If we keep the first $\tau C$ coefficients ($\tau<1$), the essential information of $\mathbf{X}$ is still preserved, while the input and output dimensions of the linear layers ($\mathcal{L}_Q, \mathcal{L}_K, $ and $\mathcal{L}_V$) are reduced from $\mathbb{R}^{N\times M^2 \times C}$ to $\mathbb{R}^{N\times M^2 \times \tau C}$. In this way, the computational cost from $\mathcal{L}_Q, \mathcal{L}_K$, and $\mathcal{L}_V$ is reduced significantly. In our experiments, $\tau $ is chosen from 25\%, 50\%, and 75\%. After obtaining the attention $\mathbf{A}\in \mathbb{R}^{N\times M^2 \times \tau C}$ from the queries $\mathbf{Q}$, keys $\mathbf{K}$, and values $\mathbf{V}$, we apply zero-padding along the last channel to make the dimension back to $\mathbb{R}^{N\times M^2 \times C}$. Before being fed into the linear layer $\mathcal{L}_O$, these padded patches are decoded by the inverse DCT (IDCT):
\begin{equation}
    \resizebox{.91\linewidth}{!}{$
            \displaystyle
     \{\mathbf{A}(\mathcal{D}^{-1})^T\}[n] = \frac{\sqrt{2}}{2}\mathbf{A}[0]+\sum_{k=1}^{C-1}\mathbf{A}[k]\cos\frac{(2n+1)k\pi}{2C}, n=0, 1,\ldots, C-1.
    \label{eq: IDCT}
            $}.
\end{equation}
Because $\mathcal{D}$ is orthogonal, \textit{i.e.} $\mathcal{DD}^T=\mathbf{I}_{C\times C}$, the IDCT matrix can be obtained directly from the transpose of the DCT matrix.
Although DCT and IDCT can be implemented in a fast manner with the complexity of $O(C\log_2 C)$ using the butterfly algorithm~\cite{chen1977fast}, such an implementation is not officially supported in PyTorch currently. Therefore, in this work, we implement the DCT and IDCT via matrix multiplication between the input tensor and the truncated DCT and IDCT matrices $\bar{\mathcal{D}}\in\mathbb{R}^{\tau C\times C}$ and $\bar{\mathcal{D}}^{-1}\in\mathbb{R}^{C\times\tau C}$. These transform matrices are obtained by truncating the DCT and IDCT matrices along rows or columns to keep $\tau$ coefficients or zero padding, respectively. Then, DCT with truncation or IDCT with zero padding is implemented as one matrix multiplication to reduce memory pressure. Furthermore, as no non-linearity exists between IDCT and the final linear layer $\mathcal{L}_O$, these two steps can be implemented as a single linear layer whose weight matrix is $\mathbf{W}_O\bar{\mathcal{D}}^{-1}$ and the bias is still $\mathbf{b}_O$. However, such a combination is not applied on DCT and $\mathbf{W}_Q$, $\mathbf{W}_K$, or $\mathbf{W}_V$. This is because it may not always reduce the computational cost for them: To compute $\mathbf{Q}$, $\mathbf{K}$, and $\mathbf{V}$, with such a combination, $3kNM^2C^3$ multiplications are needed. In contrast, if they are computed separately, there are $\tau NM^2C^3$ multiplications from DCT and $\tau^2 NM^2C^3$ multiplications from each linear layer. As a result, there are a total of $(\tau+3\tau^2)NM^2C^3$ multiplications. Hence, combing DCT with $\mathbf{W}_Q$, $\mathbf{W}_K$ and $\mathbf{W}_V$ brings computational cost saving only when $\tau>\frac{2}{3}$. To maintain the consistency of the framework, we do not combine DCT with $\mathbf{W}_Q$, $\mathbf{W}_K$, or $\mathbf{W}_V$.


\begin{table}[tb]
    \centering
  \small
    \begin{tabular}{lcc}
    \toprule
        Operation&Vanilla MSA&DCT-Compressed MSA\\
        \midrule
        DCT&-&$\tau NM^2C^3$\\
        $\mathcal{L}_{Q/K/V}$&$3NM^2C^3$&$3\tau^2NM^2C^3$\\
        $\frac{\mathbf{QK}^T}{\sqrt{d}}+\mathbf{B}$&$NM^4C^2$&$\tau^2NM^4C^2$\\
        Multiply $\mathbf{V}$&$NM^4C^2$&$\tau^2NM^4C^2$\\
        $\mathcal{L}_O$&$NM^2C^3$&$\tau NM^2C^3$\\
        \midrule
        \multirow{2}{*}{Total}&$4NM^2C^3$&($2\tau$+$3\tau^2)NM^2C^3$\\
        &+$2NM^4C^2$&+$2\tau^2NM^4C^2$\\
        \bottomrule
\end{tabular}
    \caption{Multiplications in MSA on $\mathbf{X}\in\mathbb{R}^{N\times M^2 \times C}$. $\tau<1$.}
    \label{tab:flops}
\end{table}

Computational cost comparison of the vanilla versus the proposed DCT-compressed attentions is presented in Table~\ref{tab:flops}. We evaluate both naive and simplified implementations of the DCT-compressed attention with the vanilla attention. The simplification is implemented by fusing IDCT into the linear layer $\mathcal{L}_O$. Computational overhead from the Softmax function is omitted because the Softmax function is based on exponential terms. Despite this, the Softmax function in the DCT-compressed attention requires less computational cost than in the vanilla attention as the number of entries in $\left(\frac{\mathbf{QK}^T}{\sqrt{d}}+\mathbf{B}\right)$ in the DCT-compressed attention is $\tau^2$ times ($\tau<1$) as in the vanilla attention.

\subsection{Frequency Domain Motivation of the DCT-based Initialization}
\begin{figure*}[tb]
\centering
\begin{minipage}{0.77\linewidth}
    \centering
    \subfloat{\includegraphics[width=0.24\linewidth]{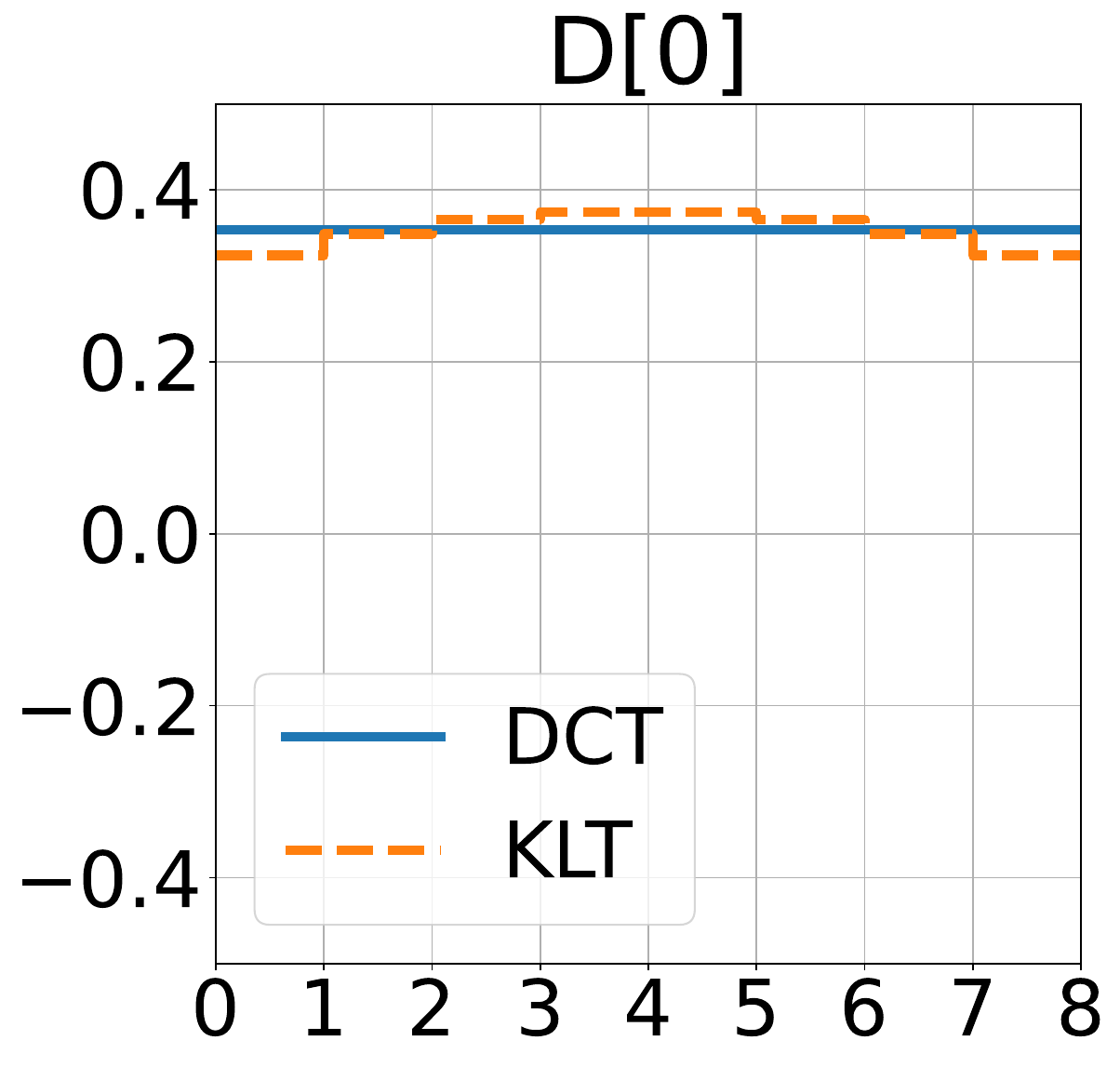}}
    \subfloat{\includegraphics[width=0.24\linewidth]{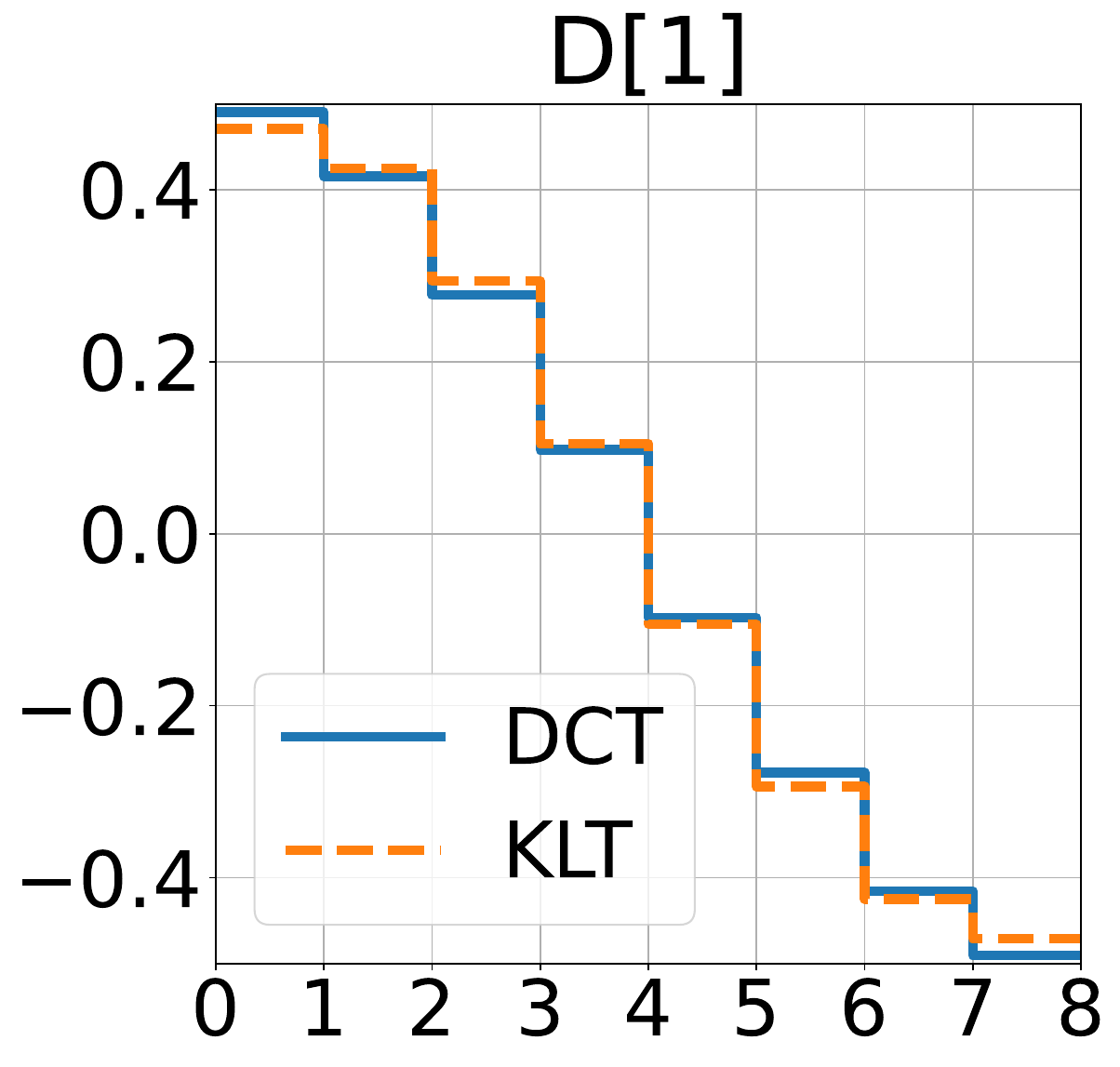}}
    \subfloat{\includegraphics[width=0.24\linewidth]{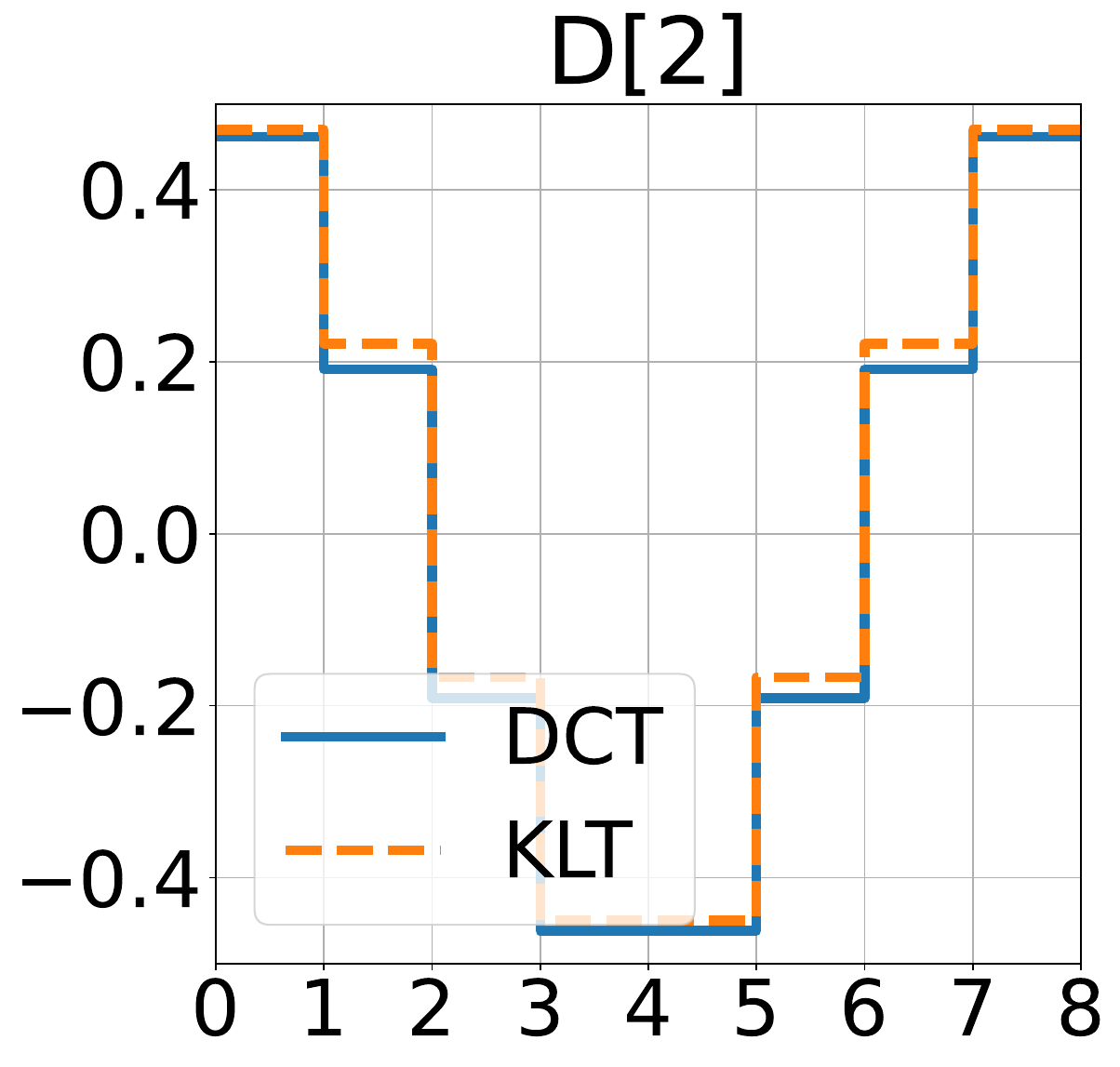}}
    \subfloat{\includegraphics[width=0.24\linewidth]{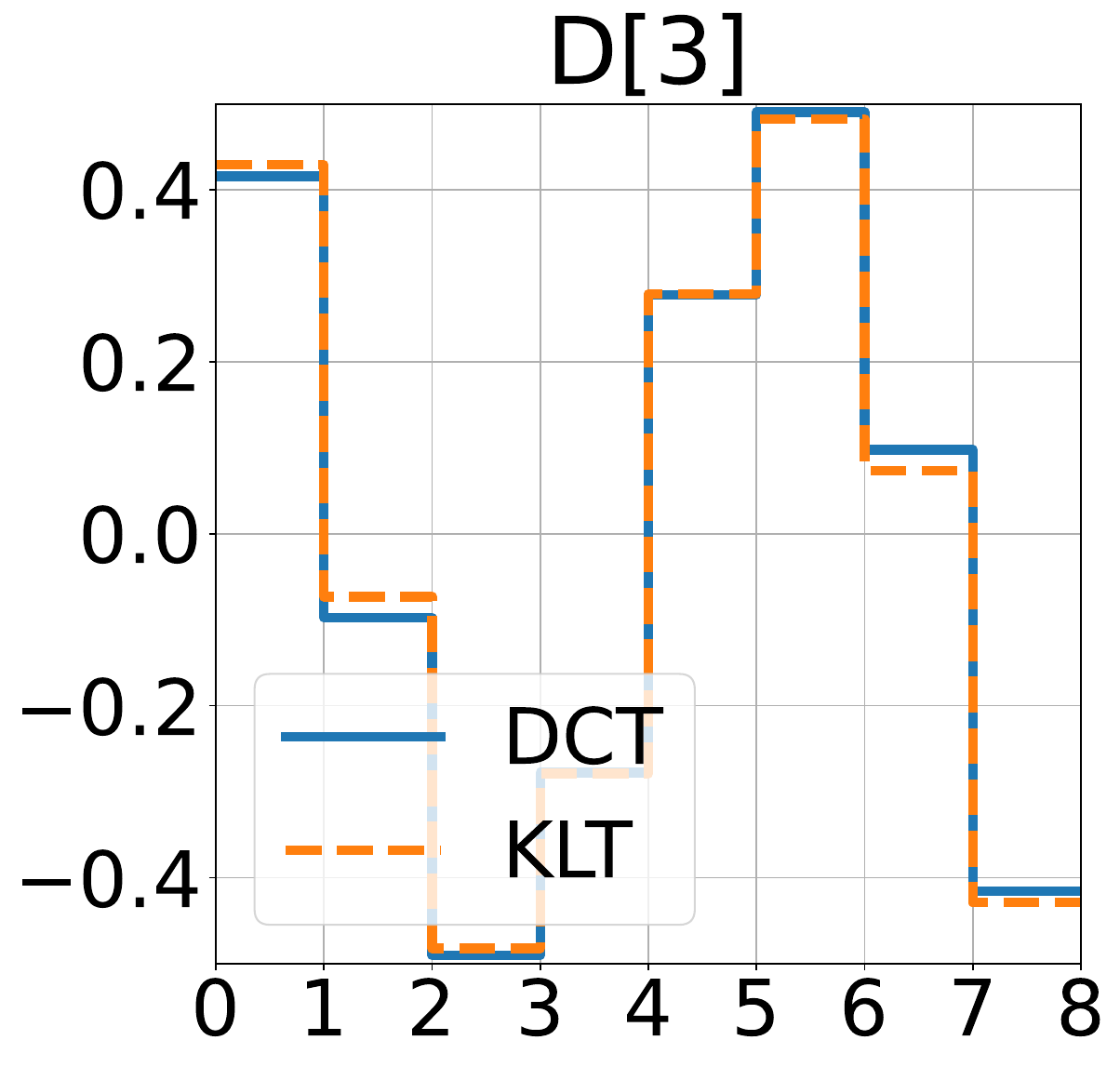}} \\
    \subfloat{\includegraphics[width=0.24\linewidth]{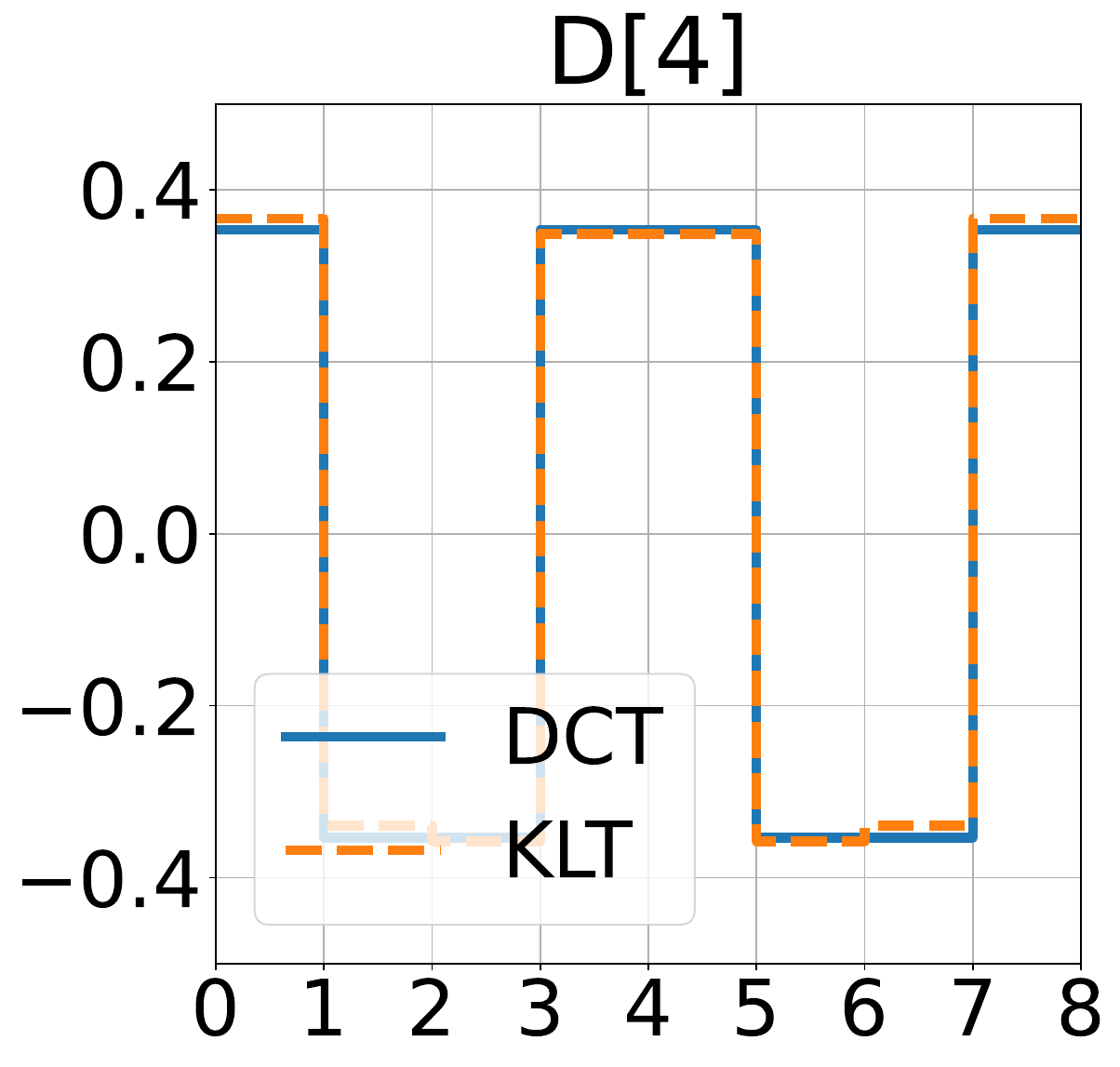}}
    \subfloat{\includegraphics[width=0.24\linewidth]{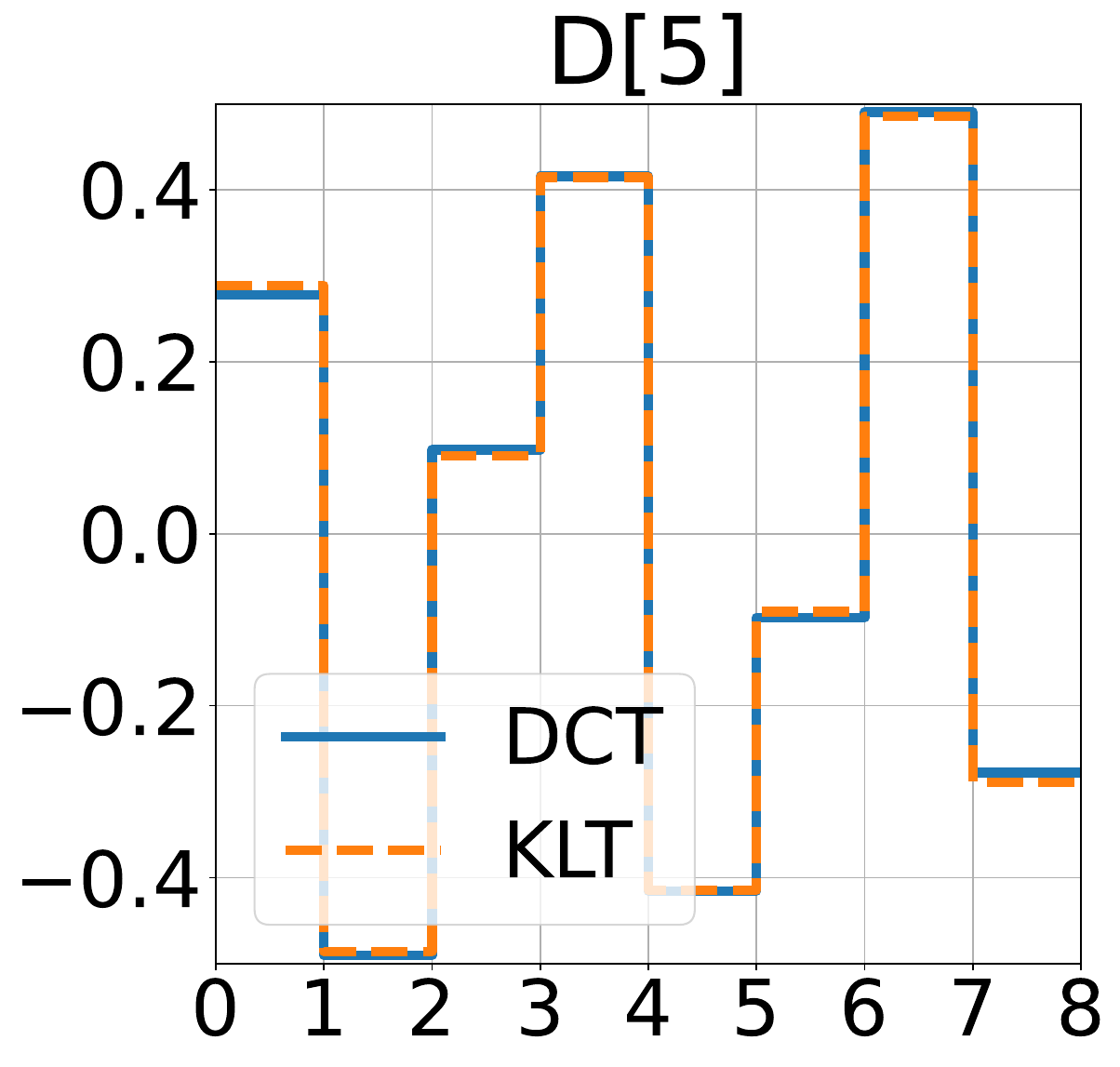}}
    \subfloat{\includegraphics[width=0.24\linewidth]{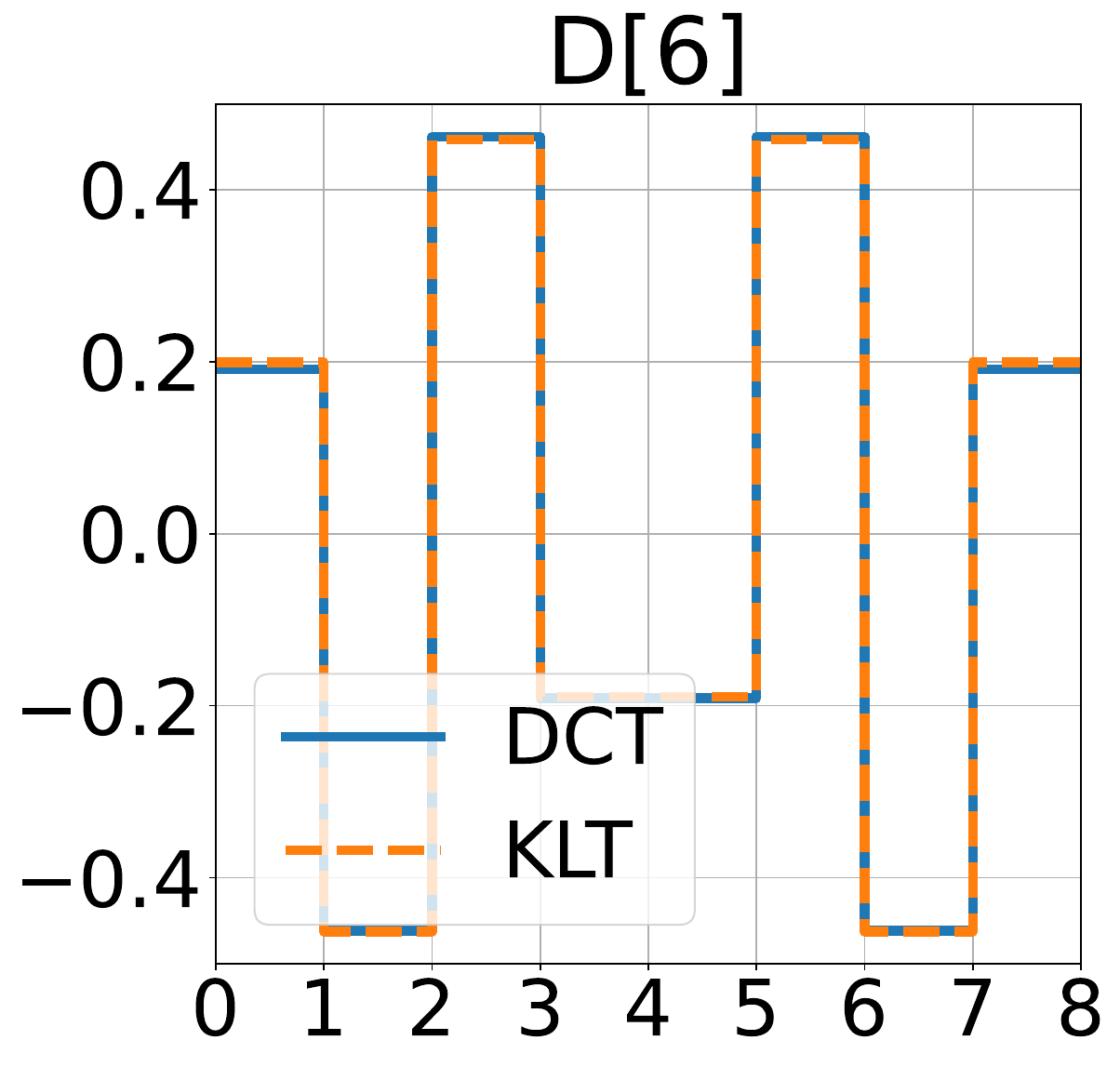}}
    \subfloat{\includegraphics[width=0.24\linewidth]{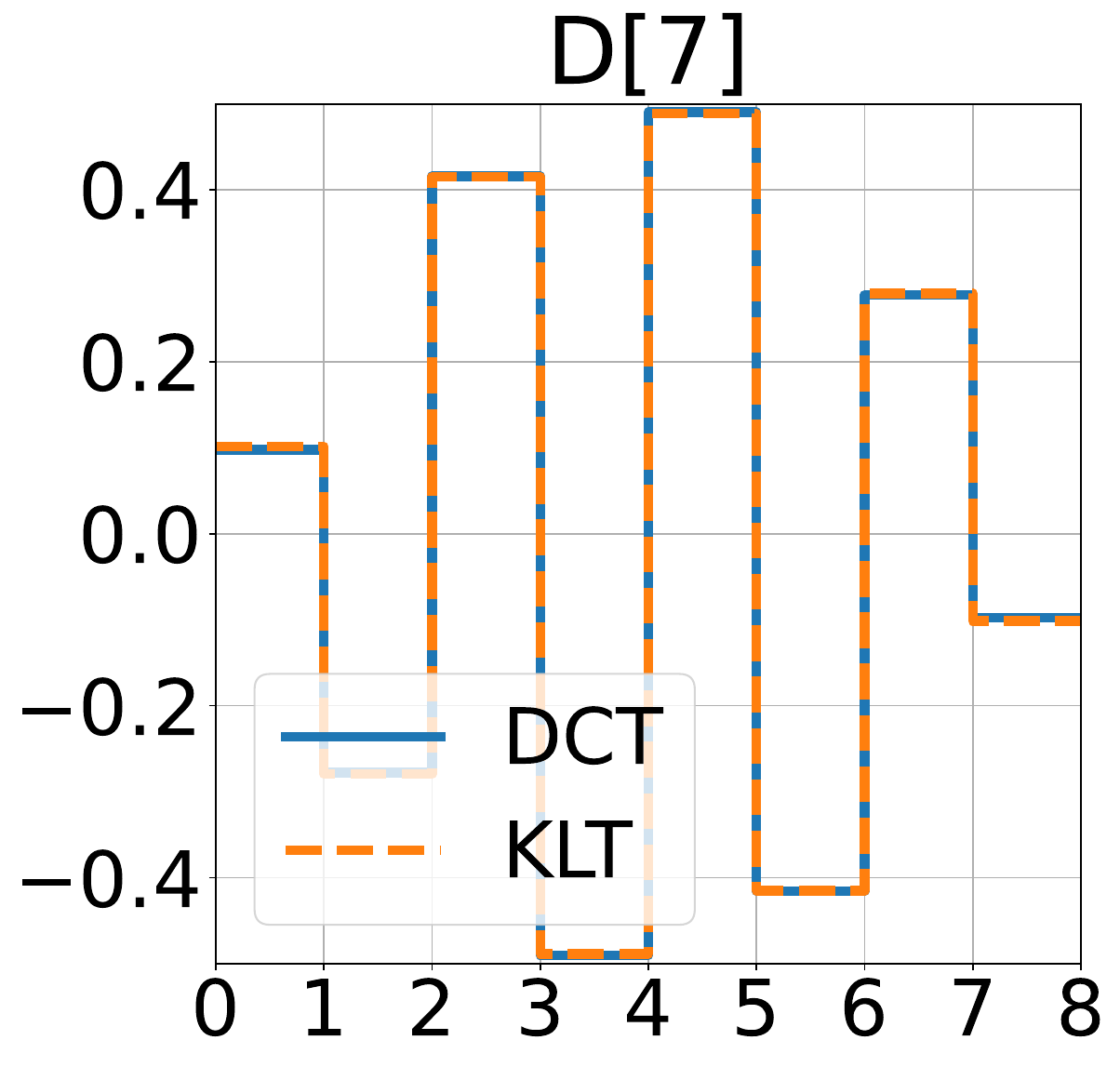}}\\
\end{minipage}
\hfill
\begin{minipage}{0.22\linewidth}
    \centering
    \subfloat[Frequency response. Each frequency response is shown in a solid curve in different colors, and their squared sum is denoted as the black dashed line denotes. \label{fig:F}]{\includegraphics[width=\linewidth]{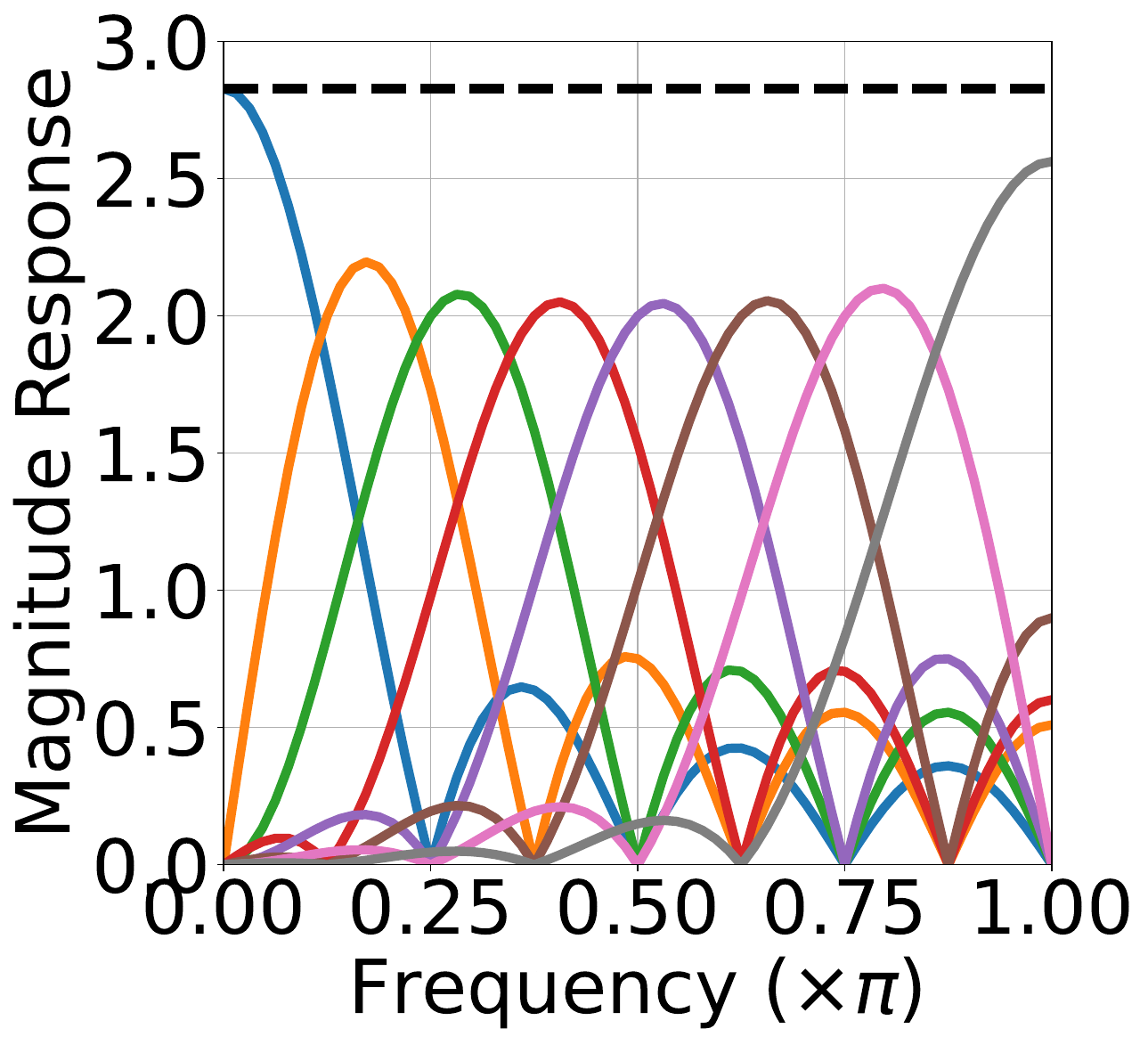}}
\end{minipage}

\caption{Basis vectors of an $8\times 8$ DCT matrix and their frequency response. The DCT basis vectors provide a good approximation to the eigenvectors of the Toeplitz matrix with $\rho=0.9$ (KLT) and cover the entire frequency spectrum.}
\label{fig:DFT(D)}
\end{figure*}

In traditional random initialization, all the weight matrices are initialized as random numbers that are independently generated. As a result, the initial matrices can be considered as a realization of white noise covering the entire spectrum. A good initialization should take advantage of the full bandwidth to cover the entire spectrum of the input. The DCT-based initialization covers the entire frequency domain:
The set of DCT basis vectors $\{\frac{\sqrt{2}}{2}, \cos\frac{(2n+1)k\pi}{2C}\}$ is the class of discrete Chebyshev polynomials:
\begin{equation}
    T_k[n] = \begin{cases}\frac{\sqrt{2}}{2}, &k=0,\\
    \cos\frac{(2n+1)k\pi}{2C}, &k=1,\ldots, C-1,\end{cases}\label{eq:chebyshev}
\end{equation}
for $n=0, 1,\ldots, C-1$. The basis member $\cos\frac{(2n+1)k\pi}{2nC}$ is the $k$-th Chebyshev polynomial $T_k[\xi]$ evaluated at the $n$-th zero of $T_n[\xi]$~\cite{ahmed1974discrete}. Fig.~\ref{fig:DFT(D)} draws the basis vectors in Eq. (\ref{eq:chebyshev}) for an $8\times8$ DCT matrix and the magnitude of their discrete Fourier transform ($\mathcal{F}\{\cdot\}$):
\begin{equation}
    \hat{T_l}[k] = \mathcal{F}\{T_l[n]\}, l=0, 1, \ldots, C-1.
\end{equation}
The entire spectrum is covered as $\sqrt{\sum_{l=0}^{C-1}|\hat{T_l}[k]|^2}$ has perfect full-band coverage, and each weight vector in the DCT-based initialization covers a different bandwidth. On the other hand, in the traditional random initialization, all of the weights start with the same spectra. This is why our DCT-based initialization is superior to the traditional random initialization for the attention layer.


\subsection{Motivation of the DCT-based Compression}
The mechanism behind the DCT-based compression mimics the Karhunen-Loeve transform (KLT), which is also known as principal component analysis (PCA), a mathematical technique used for dimensionality reduction in signal processing
~\cite{ji2022colorformer}. Neighboring image patches are highly correlated with each other. During training, the attention matrices are inter-correlated because they are obtained using similar gradient vectors. Therefore, they can be decorrelated and compressed using DCT. 
When the input data is highly correlated, the covariance matrix can be approximated by a Teoplitz matrix $\phi$ of $[1, \rho, \rho^2, \cdots, \rho^{C-1}]$ 
with $\rho<1$ close to 1~\cite{akansu2012toeplitz}. This is the covariance matrix of an autoregressive model random process AR(1) with the correlation coefficient $\rho$
~\cite{testa2016compressive}. Fig.~\ref{fig:DFT(D)} shows that DCT basis vectors approximate the KLT of a first-order Markov process with high correlation among the neighboring samples~\cite{dony2001karhunen,radunz2022low}. We observe that $\mathbf{Q}$, $\mathbf{K}$, and $\mathbf{V}$ have parameters with high correlations among themselves. We can directly use DCT in data compression since its matrix approximates the corresponding KLT matrix~\cite{ahmed1974discrete}. 
Although KLT can also be utilized in data compression, it requires massive computation for the covariance matrices and the corresponding eigenvectors of $\mathbf{X}$. On the other hand, DCT is very successful in decorrelating the data without computing the covariance matrix and eigenvectors. As a consequence, it is also used in JPEG image
~\cite{wallace1991jpeg} and MPEG video~\cite{ebrahimi2000mpeg} compression. 

\section{Experimental Results}
We conduct experiments on the CIFAR-10 and ImageNet-1K classification tasks and COCO 2017 detection and segmentation tasks. Swin Transformers~\cite{liu2021swin} were employed as the backbone network. Models were trained on a server with 8 NVIDIA A100 80GB GPUs using Python with the PyTorch framework. 

\subsection{Implementation Details}
\paragraph{Image Classification}
Training models for the image classification tasks from scratch mostly followed the settings in~\cite{liu2021swin}. We employed AdamW optimizer 
for 300 epochs using a cosine decay learning rate scheduler and 20 epochs of linear warm-up. We set the batch size as 1024, the initial learning rate as 0.001, and the weight decay as 0.05. Data augmentation was performed as follows: The CIFAR-10 augmentation followed~\cite{he2016deep}. First, the training images were padded with 4 pixels. Then, they were randomly cropped to get 32 by 32. Finally, the images were randomly flipped horizontally. The images were normalized with the means of [0.4914, 0.4822, 0.4465] and the standard deviations of [0.2023, 0.1994, 02010]. The ImageNet-1K augmentation followed PyTorch Torchvision's GitHub repository~\cite{paszke2019pytorch}. 
As a start, images were resized to make the smaller edge of the image 232 for Swin-T and 246 for Swin-S. Next, they were cropped to 224 by 224 images and randomly flipped horizontally. Techniques such as Autoaugment~\cite{cubuk2018autoaugment}, mixup~\cite{zhang2018mixup}, cutmix~\cite{yun2019cutmix}, and random erasing~\cite{zhong2020random} were applied. Images were normalized with the means of [0.485, 0.456, 0.406] and the standard deviations of [0.229, 0.224, 0.225]. During the training, the best models were stored based on the top-1 accuracy on the CIFAR-10 test dataset and the ImageNet-1K validation dataset.


\paragraph{COCO Detection and Segmentation}
Training models for the COCO tasks followed the settings in~\cite{liu2021swin}. Models started from ImageNet-1K pre-trained weights. We applied AdamW optimizer for 36 epochs with an initial learning rate of 0.0001, weight decay of 0.05, batch size of 16, 3$\times$ schedule, and multi-scale training.


\subsection{Image Classification Benchmark}

\paragraph{CIFAR-10} To demonstrate the benefits of using our DCT-based decorrelated attention on the small-scale dataset, we conduct Swin Transformer Tiny (Swin-T) from-scratch CIFAR-10 training. Table~\ref{tab:CIFAR-10} presents the performance comparison on the CIFAR-10 test dataset, where the number of parameters only counts trainable parameters. The baseline vanilla Swin-T obtained 90.51\% accuracy, and its accuracy was always improved by the DCT-based initialization regardless of which attention weight was initialized as the DCT matrix. When the weights to compute keys ($\mathbf{K}$) were initialized as the DCT matrices, the accuracy was improved to 90.83\%. For DCT-compressed attention, when we only keep 25\% of DCT coefficients, trainable parameters were reduced from 27.53M to 21.45M (22.1\% off), while the accuracy only dropped 0.05\%. If we kept more DCT coefficients, the accuracy even improved. 75\% of DCT coefficients increased accuracy from 90.51\% to 90.90\%. This is because although the number of trainable parameters is reduced, transforming the input data into the frequency domain using DCT can boost the extraction of important features. 

\begin{table}[tb]
  \centering
  \small
  \begin{tabular}{lcc}
    \toprule
    Model &Param(M) & Top-1 Acc \\
    \midrule
    Swin-T&27.53&90.51\%\\
    \rowcolor{Better} Swin-T-DCT-Q&27.53&90.78\%\\
    \rowcolor{Better} Swin-T-DCT-K&27.53&90.83\%\\
    \rowcolor{Better} Swin-T-DCT-V&27.53&90.81\%\\
    \rowcolor{Worse}Swin-T-DCT-0.25 &21.45&90.46\%\\
    \rowcolor{Better}Swin-T-DCT-0.5 &22.67&90.67\%\\
    \rowcolor{Better}Swin-T-DCT-0.75 &24.69&90.90\%\\
    \bottomrule
  \end{tabular}
  \caption{CIFAR-10 test dataset evaluation. ``-DCT-'' followed by Q/K/V in model names indicates which attention projection weights are initialized with the DCT matrix, while numerical values denote the proportion of coefficients retained in DCT-compressed attention.  Models with higher or lower top-1 accuracy are highlighted in pink or cyan, respectively.}
  \label{tab:CIFAR-10}
\end{table}


\begin{table}[tb]
  \centering
  \small
  \begin{tabular}{lcc}
    \toprule
    Method& Param/Flops& Top-1/Top-5 Acc \\
    \midrule
    Swin-T &28.29M/4.49G&81.474\%/95.776\%\\
    PoolFormer-S24&21M/3.5G&80.3\%/-\\
    DCFormer-SW-T($\tau$=1)&28.29M/4.5G&81.2\%/-\\
    DCFormer-SW-T($\tau$=2)&28.29M/1.3G&79.2\%/-\\
    gSwin-T&22/3.6&81.715\%/-\\

    \rowcolor{Better} Swin-T-DCT-Q&28.29M/4.49G&81.568\%/95.862\%\\
    \rowcolor{Better} Swin-T-DCT-K&28.29M/4.49G&81.728\%/95.862\%\\
    \rowcolor{Better} Swin-T-DCT-V&28.29M/4.49G&81.528\%/95.850\%\\
    \rowcolor{Worse} Swin-T-DCT-0.25&22.21M/3.24G&79.580\%/94.936\%\\
    \rowcolor{Worse} Swin-T-DCT-0.5&23.43M/3.64G&80.762\%/95.520\%\\
    \rowcolor{Worse} Swin-T-DCT-0.75&25.45M/4.18G&81.474\%/95.728\%\\
    \midrule
    Swin-S &49.61M/8.74G&83.196\%/96.360\%\\
    PoolFormer-M36&56M/8.8G&82.1\%/-\\
    DCFormer-SW-S($\tau$=1)&49.61M/8.7G&82.8\%/-\\
    DCFormer-SW-S($\tau$=2)&49.61M/2.7G&80.9\%/-\\
    gSwin-S&40M/7.0G&83.014\%/-\\ 
    \rowcolor{Better} Swin-S-DCT-Q&49.61M/8.74G&83.298\%/96.298\%\\
    \rowcolor{Better} Swin-S-DCT-K&49.61M/8.74G&83.378\%/ 96.398\%\\
    \rowcolor{Better} Swin-S-DCT-V&49.61M/8.74G&83.200\%/96.364\%\\
    \rowcolor{Worse} Swin-S-DCT-0.25 &38.54M/6.28G&81.840\%/95.704\%\\
    \rowcolor{Worse} Swin-S-DCT-0.5 &40.76M/7.07G&82.622\%/96.036\%\\
    \rowcolor{Better} Swin-S-DCT-0.75 &44.45M/8.13G&83.202\%/96.388\%\\
    \midrule
    ViT-B-32&88.22M/4.41G&73.120\%/90.306\%\\
    \rowcolor{Better} ViT-B-32-DCT-Q&88.22M/4.41G&73.278\%/90.574\%\\
    \rowcolor{Better} ViT-B-32-DCT-K&88.22M/4.41G&73.614\%/90.870\%\\
    \rowcolor{Better} ViT-B-32-DCT-V&88.22M/4.41G&73.306\%/90.502\%\\
    \rowcolor{Worse} ViT-B-32-DCT-0.5&66.97M/3.51G&72.948\%/90.390\%\\
    \rowcolor{Better} ViT-B-32-DCT-0.75&75.83M/4.05G&73.226\%/90.642\%\\
    \bottomrule
  \end{tabular}
  \caption{ImageNet-1K validation dataset evaluation. }
  \label{tab:ImageNet}
\end{table}

\paragraph{ImageNet-1K}
Table~\ref{tab:ImageNet} presents the performance comparison on the ImageNet-1K dataset. Same as on the CIFAR-10 dataset, DCT-based initialization consistently improves performance on both Swin-T and Swin-S models, regardless of which attention weight was initialized as the DCT matrix. For example, when we initialized the weights to compute keys ($\mathbf{K}$) as the DCT weights, the accuracy of Swin-T was improved from 81.474\% to 81.728\%, and Swin-S was increased from 83.196\% to 83.378\%, respectively. We then evaluated the DCT-compressed attention with different truncating rates and compared the results with DCFormer~\cite{li2023discrete}, a DCT-based approach designed to reduce computational cost. When 75\% of DCT coefficients were kept, we obtained a comparable accuracy result on Swin-T and a slightly better accuracy result on Swin-S, while reducing the trainable parameters from 28.29M to 25.45M (13.1\% off) and from 49.61M to 44.45M (10.4\% off), respectively. Flops were reduced from 4.49G to 4.18G (6.9\% off) for Swin-T and 8.13G to 8.74G (7.0\% off) for Swin-S. Retaining fewer coefficients yields further savings at the cost of accuracy.  In contrast, DCFormer-SW-T/S with $\tau=2$ ($4\times$ frequency compression) substantially reduces FLOPs but suffers notable performance degradation, as it treats low- and high-frequency components equally.
We further validated our methods on the ViT-B-32 backbone~\cite{dosovitskiy2020image}. DCT-based initialization of the key projection improves accuracy from 73.120\% to 73.614\%. Applying DCT-compressed attention with 75\% coefficient retention further improves accuracy to 73.226\%, while reducing parameters from 88.22M to 75.83M (14.0\%) and FLOPs from 4.41G to 4.05G (8.2\%).

\subsection{Ablation Study}\label{sec: ablation study}
\paragraph{Non-Trainable DCT Weights Initialization}
This study investigates the effectiveness of DCT weights as a principled starting point for attention projection initialization. Specifically, we fix the projection weights as DCT matrices, under which the computation of queries, keys, and values becomes equivalent to applying fixed shifts in the DCT representation of the input features. Table~\ref{tab:non-trainable} shows that our method achieved accuracy comparable to the baseline on the CIFAR-10 dataset while requiring fewer trainable parameters. On the ImageNet-1K dataset, non-trainable DCT weights for computing $\mathbf{Q}$ or $\mathbf{K}$ led to only a marginal drop in top-1 accuracy (less than 0.3\%). In this work, the DCT was implemented via matrix multiplication, incurring the same computational cost as standard attention; however, this overhead can be further reduced using butterfly-based DCT implementations.


\begin{table}[tb]
  \centering
  \small
  \begin{tabular}{l|cc|cc}
    \toprule
    \multirow{2}{*}{DCT} & \multicolumn{2}{c|}{CIFAR-10}& \multicolumn{2}{c}{ImageNet-1K}\\
    & Param& Top-1 Acc & Param& Top-1/Top-5 Acc\\
    \midrule
    -&27.53M&90.51\%&28.29M&81.474\%/95.776\%\\
    Q&\cellcolor{Worse}25.37M&\cellcolor{Worse}90.47\%&\cellcolor{Worse}26.13M&\cellcolor{Worse}81.214\%/\cellcolor{Worse}95.674\%\\
    K&\cellcolor{Better}25.37M&\cellcolor{Better}90.64\%&\cellcolor{Worse}26.13M&\cellcolor{Worse}81.302\%/\cellcolor{Worse}95.788\%\\
    V&\cellcolor{Better}25.37M&\cellcolor{Better}90.75\%&\cellcolor{Worse}26.13M&\cellcolor{Worse}80.486\%/\cellcolor{Worse}95.236\%\\
    QK&\cellcolor{Better}23.21M&\cellcolor{Better}90.70\%&\cellcolor{Worse}23.98M&\cellcolor{Worse}80.608\%/\cellcolor{Worse}95.446\%\\
    QV&\cellcolor{Worse}23.21M&\cellcolor{Worse}90.50\%&\cellcolor{Worse}23.98M&\cellcolor{Worse}80.234\%/\cellcolor{Worse}95.208\%\\
    KV&\cellcolor{Better}23.21M&\cellcolor{Better}90.62\%&\cellcolor{Worse}23.98M&\cellcolor{Worse}80.228\%/\cellcolor{Worse}95.276\%\\
    \bottomrule
  \end{tabular}
\caption{Ablation study on Swin-T with non-trainable DCT weights.}
  \label{tab:non-trainable}
\end{table}


\paragraph{Multiple DCT Initializations}
This study shows why we only initialize one instead of multiple projection weights. Table~\ref{tab:ImageNet multi} presents an ablation study on initializing multiple projection matrices with the DCT matrix. Although this approach still improved top-1 accuracy for Swin-T, the gains were reduced compared to initializing only a single matrix. For example, initializing both $\mathbf{W}_Q$ and $\mathbf{W}_K$ resulted in a top-1 accuracy of 81.540\%, lower than 81.568\% by initializing only $\mathbf{W}_Q$ and 81.728\% by initializing only $\mathbf{W}_K$. This degradation was likely due to the weight symmetry problem~\cite{hu2019exploring}, which often arises when multiple weight matrices are initialized with identical values. In the context of attention mechanisms, the query, key, and value projections play distinct roles in capturing relationships across the input sequence. If they are overly similar due to identical initialization, the model's ability to attend to relevant and diverse information is compromised. One of the key strengths of self-attention lies in its capacity to model complex dependencies between different positions in the input. Therefore, when $\mathbf{W}_Q$, $\mathbf{W}_K$, and $\mathbf{W}_V$ share the same initialization, this capacity is diminished, leading to reduced performance. We further verified this observation using Swin-S by applying the DCT-based initialization to both $\mathbf{W}_Q$ and $\mathbf{W}_K$. The results confirmed the same trend, supporting our conclusion that single-matrix DCT initialization is generally more effective.


\begin{table}[tb]
  \centering
  \begin{tabular}{lcc}
    \toprule
    Model& Param/Flops& Top-1/Top-5 Acc\\
    \midrule
    Swin-T&28.29M/4.49G&81.474\%/95.776\%\\
    \rowcolor{Better} Swin-T-DCT-Q&28.29M/4.49G&81.568\%/95.862\%\\
    \rowcolor{Better} Swin-T-DCT-K&28.29M/4.49G&81.728\%/95.862\%\\
    \rowcolor{Better} Swin-T-DCT-V&28.29M/4.49G&81.528\%/95.850\%\\
    \rowcolor{Better} Swin-T-DCT-QK&28.29M/4.49G&81.540\%/95.754\%\\
    \rowcolor{Better} Swin-T-DCT-QV&28.29M/4.49G&81.548\%/95.748\%\\
    \rowcolor{Better} Swin-T-DCT-KV&28.29M/4.49G&81.536\%/95.770\%\\
    \midrule
     Swin-S&49.61M/8.74G&83.196\%/96.360\%\\
    \rowcolor{Better} Swin-S-DCT-Q&49.61M/8.74G&83.298\%/96.298\%\\
    \rowcolor{Better} Swin-S-DCT-K&49.61M/8.74G&83.378\%/96.398\%\\
    \rowcolor{Better} Swin-S-DCT-QK&49.61M/8.74G&83.208\%/96.378\%\\
    \bottomrule
  \end{tabular}
\caption{Ablation study on ImageNet-1K for multiple weights initialization as the DCT matrix.}
  \label{tab:ImageNet multi}
\end{table}

\begin{figure}[tb]
\centering
{\includegraphics[width=1\linewidth]{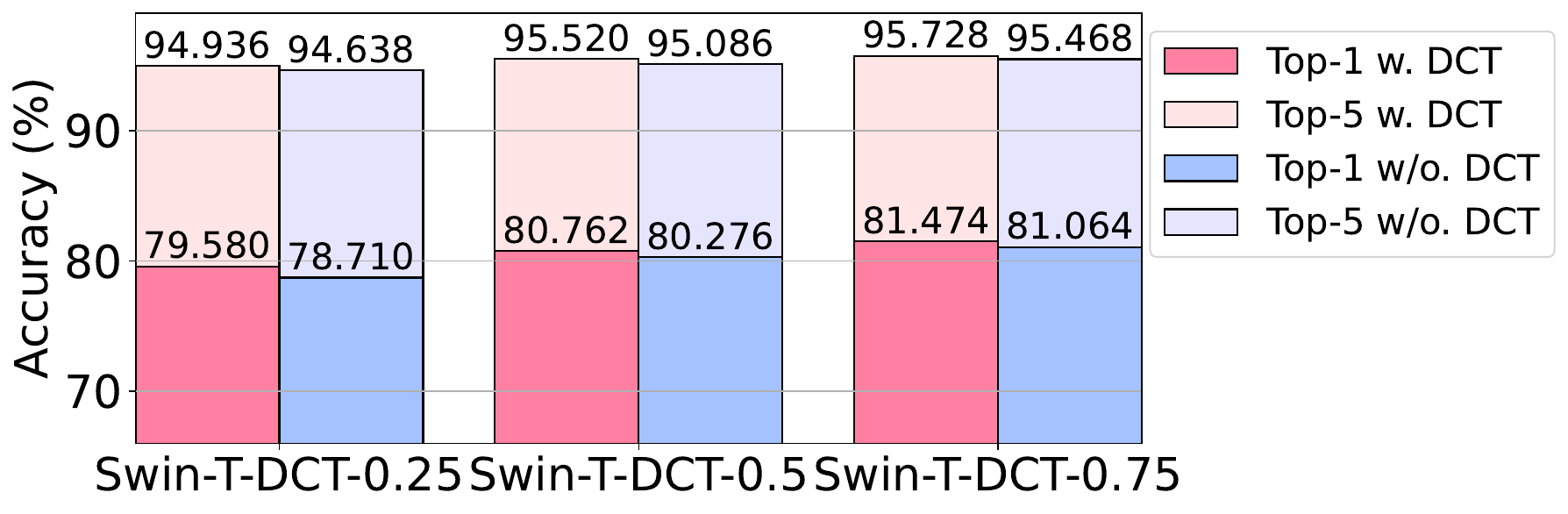}}
\caption{Ablation study on ImageNet-1K for removing DCT from DCT-compressed attention. Swin-T serves as the backbone model.}
\label{fig:nodct}
\end{figure}

\paragraph{Removing DCT in Compressed Attention}
This study discusses the necessity of DCT in the DCT-compressed attention. Specifically, removing DCT reduces the compression mechanism to a form of direct coefficient pruning in the spatial domain, without exploiting frequency-domain decorrelation. As Fig.~\ref{fig:nodct} presents, when we removed the DCT in the DCT-compressed attention and retained other components, the accuracy of the revised Swin-T models on ImageNet-1K decreased across all compression ratios. These experiments show that DCT is essential because it transforms feature representations into the frequency domain, reducing redundancy and enhancing the model’s ability to capture essential information. As a result, the attention mechanism benefits from improved expressiveness and robustness, even under compression. The accuracy drop observed in Fig.~\ref{fig:nodct} confirms that DCT plays a crucial role in preserving discriminative features, demonstrating its necessity in the proposed approach.

\begin{table}[tb]
  \centering
  \small
  \begin{tabular}{lc|ccc|ccc}
    \toprule
    \multirow{2}{*}{Backbone}&\multirow{2}{*}{Param}&Object& Instance\\
    &&Detection&Segmentation\\
    \midrule
    Swin-T&86M&50.3/69.1/54.3&43.7/ 66.6/47.3\\
    DCFormer-T&86M&43.5/62.6/47.4&38.0/59.3/40.7\\
    Swin-T-DCT-K&86M&\cellcolor{Better}50.4/\cellcolor{Better}69.2/\cellcolor{Better}54.3&\cellcolor{Better}43.9/\cellcolor{Better}66.6/\cellcolor{Better}47.6\\
    Swin-T-DCT-0.75&83M&\cellcolor{Worse}50.0/\cellcolor{Worse}69.1/\cellcolor{Worse}53.9&\cellcolor{Better}43.7/\cellcolor{Better}66.8/\cellcolor{Better}47.6\\
    \midrule
     Swin-S&107M&51.8/70.4/56.3&44.7/67.9/ 48.5\\
     DCFormer-S&107M&46.6/64.9/50.4&40.5/62.2/43.7\\
     Swin-S-DCT-K&107M&\cellcolor{Better}52.4/\cellcolor{Better}70.8/\cellcolor{Better}57.3&\cellcolor{Better}45.1/\cellcolor{Better}68.3/\cellcolor{Better}49.1\\
     Swin-S-DCT-0.75&102M&\cellcolor{Better}52.1/\cellcolor{Better}70.6/\cellcolor{Better}57.0&\cellcolor{Better}45.0/\cellcolor{Better}68.0/\cellcolor{Better}48.8\\
    \bottomrule
  \end{tabular}
\caption{COCO results. Results are $\text{AP}^{\text{Mask}}$/ $\text{AP}^{\text{Mask}}_{0.5}$/$\text{AP}^{\text{Mask}}_{0.75}$.}
  \label{tab:coco}
\end{table}

\subsection{COCO 2017 Benchmark}
We further evaluated the proposed methods on COCO 2017 object detection and instance segmentation tasks~\cite{lin2014microsoft}. We took Swin Transformers with ImageNet-1K pre-trained weights as the backbone and the Cascade Mask R-CNN
~\cite{cai2018cascade} as the pipeline. Based on the superior performance observed in the ImageNet-1K experiments when initializing $\mathbf{W}_K$ with the DCT matrix, we adopted this initialization for the Swin-T and Swin-S backbones in our evaluation. As shown in Table~\ref{tab:coco}, both models achieved slightly higher average precision (AP) scores compared to their vanilla counterparts across both tasks. We also evaluated the DCT-compressed attention variant with the 0.75 compression ratio. This modification allowed Swin-T to maintain comparable AP performance while reducing the parameter count by approximately 3 million. It also enabled Swin-S to achieve improved AP scores with around 5 million fewer parameters.


\section{Conclusion}
In this work, we introduced two new methods using DCT to decorrelate the input in the attention for vision transformers. First, we presented a novel DCT-based initialization approach to address the challenges associated with training attention weights. We demonstrated a significant improvement over traditional initialization, providing a robust foundation for the attention mechanism. Second, we formulated a DCT-based compression method that can effectively reduce the computational overhead by truncating high-frequency components. It resulted in smaller weight matrices that compromised considerable accuracy. We validated the efficacy of these breakthroughs, showcasing their potential to enhance the performance and efficiency of Transformer models in classification tasks. These contributions underscore the importance of thoughtful initialization strategies and compression techniques in advancing the capabilities of the Vision Transformer architectures.

\section*{Acknowledgments}
 H. Pan and U. Bagci were supported by NIH grants R01-HL171376 and U01-CA268808. E. Hamdan, X. Zhu, and A. E. Cetin were supported by NSF 2531376.

\bibliographystyle{named}
\bibliography{ijcai26}
\end{document}